\documentclass[final]{cvpr}

\usepackage{times}
\usepackage{epsfig}
\usepackage{graphicx}
\usepackage{amsmath}
\usepackage{amssymb}

\usepackage{amsmath,bm}
\usepackage{amsfonts}
\usepackage{amstext}
\usepackage[american]{babel}
\usepackage{microtype}
\usepackage{booktabs}
\usepackage{subfigure}
\usepackage{multirow}
\usepackage[ruled, vlined]{algorithm2e}
\usepackage{color}

\usepackage[pagebackref=false,breaklinks=true,colorlinks,bookmarks=false]{hyperref}

\def\cvprPaperID{5222} 
\def\confYear{CVPR 2021}

\begin{document}

\title{Zero-shot Adversarial Quantization}

\renewcommand{\thefootnote}{\fnsymbol{footnote}}
\author {
  Yuang Liu~,\quad Wei Zhang\footnotemark[1]~, \quad Jun Wang\footnotemark[1] \\
  East China Normal University, Shanghai, China \\
  {\tt\small \{frankliu624, zhangwei.thu2011, wongjun\}@gmail.com}
}

\maketitle

\footnotetext[1]{Corresponding author.}
\renewcommand{\thefootnote}{\arabic{footnote}}

\begin{abstract}
Model quantization is a promising approach to compress deep neural networks and accelerate inference, making it possible to be deployed on mobile and edge devices. To retain the high performance of full-precision models, most existing quantization methods focus on fine-tuning quantized model by assuming training datasets are accessible. 
However, this assumption sometimes is not satisfied in real situations due to data privacy and security issues, thereby making these quantization methods not applicable.
To achieve zero-short model quantization without accessing training data, a tiny number of quantization methods adopt either post-training quantization or batch normalization statistics-guided data generation for fine-tuning.
However, both of them inevitably suffer from low performance, since the former is a little too empirical and lacks training support for ultra-low precision quantization, while the latter could not fully restore the peculiarities of original data and is often low efficient for diverse data generation.
To address the above issues, we propose a zero-shot adversarial quantization (ZAQ) framework, facilitating effective discrepancy estimation and knowledge transfer from a full-precision model to its quantized model.
This is achieved by a novel two-level discrepancy modeling to drive a generator to synthesize informative and diverse data examples to optimize the quantized model in an adversarial learning fashion.
We conduct extensive experiments on three fundamental vision tasks, demonstrating the superiority of ZAQ over the strong zero-shot baselines and validating the effectiveness of its main components.
Code is available at \href{https://git.io/Jqc0y}{https://git.io/Jqc0y}.

\end{abstract}

\section{Introduction}

Although deep neural networks (DNNs), especially deep convolutional networks (DCNs), have achieved remarkable performance in a broad range of computer vision tasks~\cite{krizhevsky2012imagenet,toshev2014deeppose,long2015fully,ren2017faster}, their ever-growing complexities --- a large number of model parameters --- inhibit the applications on cloud and edge devices.
As a consequence, model quantization, converting high-precision parameters to low-precision ones, becomes one of the main paradigms in model compression and acceleration~\cite{DengLHSX20}. 
To mitigate the performance degradation issue due to model quantization, quantization-aware fine-tuning approaches have been extensively studied to optimize quantized models on the full training datasets~\cite{wu2016quantized,jacob2018quantization,shen2020q}.
However, in real situations, original training data is sometimes inaccessible due to privacy and security issues.
For instance, electronic health records usually contain patients' private information.
As such, the quantization-aware fine-tuning methods are no longer applicable. 

\begin{figure}[tbp]
  \centering
  \includegraphics[width=\linewidth]{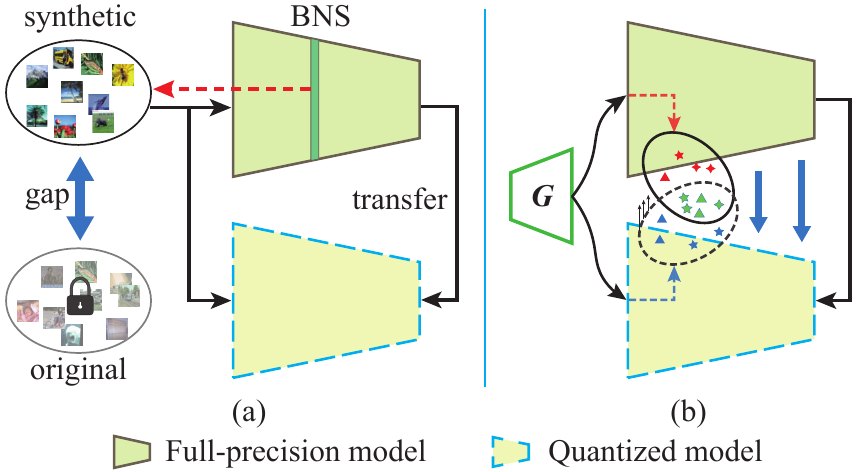}
  \caption{Overview of our framework. The methods based on sample reconstruction are shown as part (a), and part (b) is the overview of our framework. BNS is short for batch normalization statistics stored in the BN layers.}
  \label{fig:overview}
\end{figure}

Post-training quantization methods~\cite{banner2018aciq,nagel2019data,zhao2019improving} therefore emerge to quantize weights and activations in DNNs through correction strategies, without fine-tuning.
However, there is a negligible gap between the strategies and the goals of target tasks, causing the quantized models to suffer from performance degradation.
This issue is even amplified for the ultra-low precision situation. 
To address this, batch normalization statistics~(BNS)-guided data generation is leveraged by recent methods~\cite{cai2020zeroq,xu2020generative}.
They aim at synthesizing data samples that match the real-data statistics encoded in the batch normalization layers of full-precision deep models.
The synthetic data is further leveraged to fine-tune the quantized models by directly optimizing on target tasks supervised by its full-precision model, as shown in Figure~\ref{fig:overview}(a).
Although the performance of ultra-low precision models is boosted to some extent, thanks to fine-tuning, data generated by batch normalization statistics is hard to fully recover the peculiarities of training data and the generation process itself is time-consuming due to data redundancy.
These issues make the results still far from satisfactory. 

This paper seeks to promote the development of data-free model quantization by addressing the above-mentioned issues.
We, therefore, present a novel learning framework named Zero-shot Adversarial Quantization (ZAQ) to perform model quantization without utilizing any sample from training data.
Specifically, we devise a two-level discrepancy modeling strategy for ZAQ to measure the gap between a quantized model and its corresponding full-precision model.
We consider not only the output discrepancy from models' top layers, just similar as existing data-free model quantization methods, but also fuses a new intermediate inter-channel discrepancy based on feature maps.
A generator in ZAQ is responsible for generating informative and diverse data examples in an adversarial learning manner~\cite{goodfellow2014generative} --- optimization based on a minimax game --- to enable effective discrepancy estimation and knowledge transfer, as depicted Figure~\ref{fig:overview}(b).
In addition, activation regularization is adopted to facilitate the generator to obtain examples more sensitive to the network.
To sum up, our contributions are as follows:
\begin{itemize}
  \item We propose a zero-shot adversarial quantization framework to support effective data generation and knowledge transfer.
  To our best knowledge, it represents the first effort to apply adversarial learning to data-free model quantization.
  
  \item A novel two-level discrepancy modeling strategy is devised to measure the discrepancy between a quantized model and its full-precision model, thereby guiding the training of the quantized model and generator.
  
  \item We conduct extensive experiments on image classification, segmentation, and object detection tasks, showing our ZAQ framework achieves state-of-the-art results in data-free situation, works well for ultra-low precision scenarios, and is efficient compared to the approaches of BNS-guided data generation for model quantization.

\end{itemize}

\section{Related Work}


\textbf{Model quantization} is a promising model compression methods aiming to store parameters with fewer bits so that computation can be executed on integer-arithmetic units rather than on power-hungry floating-point ones~\cite{jacob2018quantization}. An important challenge with quantization is that it can lead to significant performance degradation, especially in ultra-low precision settings. To cope with this, PACT~\cite{choi2018pact} used an activation clipping parameter to find the right quantization scale. Zhu~\etal~\cite{zhu2020towards} built a flexible and unified INT8 training framework for vision tasks. Flexpoint~\cite{koster2017flexpoint}, MPT~\cite{micikevicius2017mixed} and DFP~\cite{das2018mixed} all use 16-bit floating-point to train DNNs with accuracy comparable to full-precision model. 
And there are some approaches to decrease induced degradation by quantization-aware training~\cite{banner2018scalable,jacob2018quantization,mckinstry2018discovering} or reducing the dynamic range of activations by clipping outliers~\cite{zhao2019improving,migacz20178,banner2018aciq}. Instead of focusing on improving the quantization process itself, \cite{meller2019same} explored an equivalent weight arrangement that make the net less sensitive to quantization.
However, all above quantization methods generally require access to the entire training data which is not always available as aforementioned. 

\textbf{Data-free model compression} has been a hot topic and draw more and more attention in recent years, which is a challenge to compress model without training data. Srinivas and Babu~\cite{srinivas2015data}, the pioneers in data-free compression, introduced a channel pruning method without original training data. 
Since then, more and more kinds of data-free or zero-shot compression methods were proposed, including quantization~\cite{banner2018aciq,cai2020zeroq,xu2020generative}, weight factorization~\cite{nagel2019data} and knowledge distillation (KD)~\cite{lopes2017data,chen2019data,fang2020data,ye2020data, liu2020adaptive}. 
DFQ~\cite{nagel2019data} and ACIQ~\cite{banner2018aciq} are both post-training quantization methods relying on weight equalization or bias correction without fine-tuning on the entire dataset. But when applied to ultra-low precision (\ie, lower than 6-bit) model, these kinds of quantization methods cannot prevent quantization models from performance degradation. 
Most of the data-free KD methods attempt to reconstruct the original data from pre-trained teacher model utilizing prior information about the underlying data distribution, such as BNS~\cite{yin2020dreaming}, Dirichlet distribution~\cite{nayak2019zero} and category information~\cite{chen2019data}. However, they ignore the intermediate features to guide the student network learning. 

Two recent data-free quantization studies~\cite{cai2020zeroq,xu2020generative} quantize and fine-tune models without needing original data. 
Their core idea is to reconstruct some samples from full-precision models to fine-tune quantized models. 
To be specific, ZeroQ~\cite{cai2020zeroq} directly reconstructs samples by optimizing from random noises according to BNS of full-precision models.
GDFQ~\cite{xu2020generative} further adopts a generator to reconstruct samples guided by BNS and extra category label information, which limits its application to classification tasks. 
To sum up, there is still a large gap between the data generated based on BNS and original training data after a time-consuming generation process.
Moreover, both ZeroQ and GDFQ poorly support high-level vision tasks due to the lack of considering information from intermediate layers of full-precision models.


\section{The Computational Framework}

\begin{figure}[!t]
  \centering
  \includegraphics[width=\linewidth]{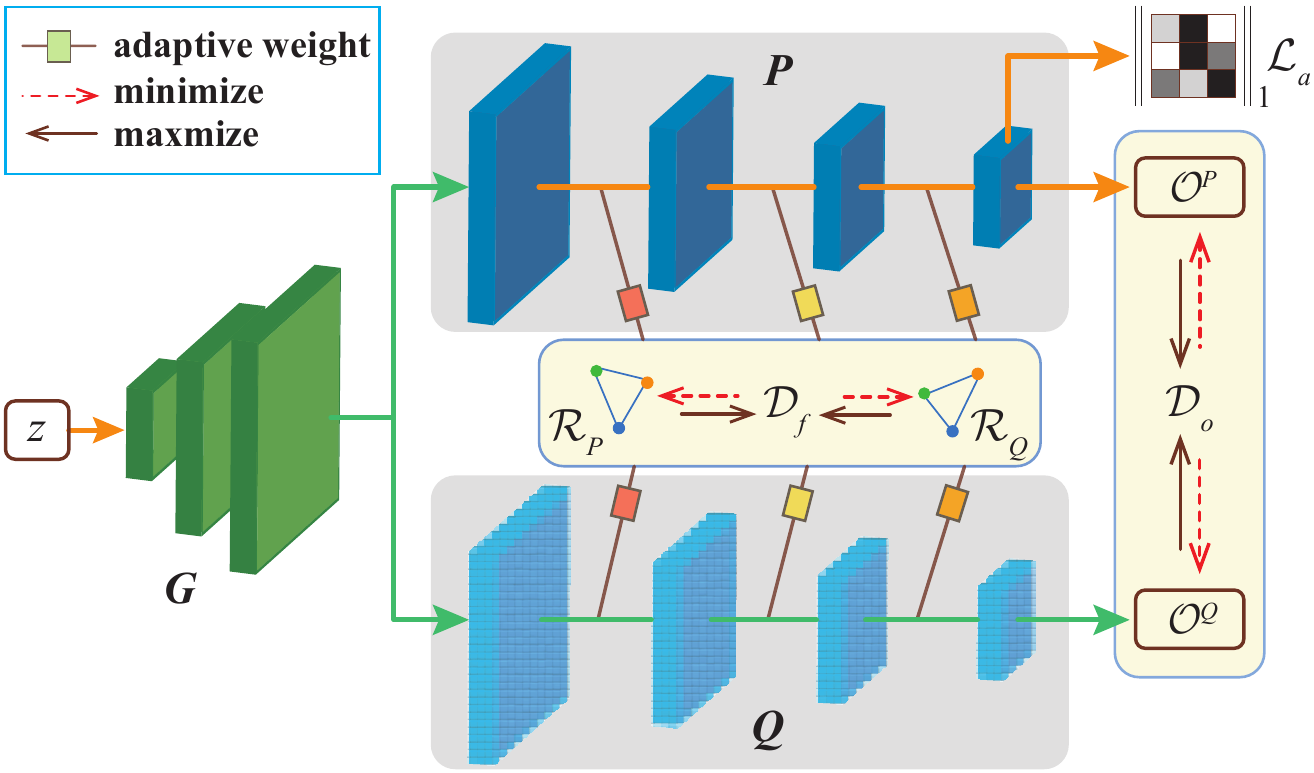}
  \caption{Framework of ZAQ.}
  \label{fig:ZAQ}
\end{figure}

\textbf{Framework Overview:} Figure~\ref{fig:ZAQ} depicts the basic framework of ZAQ. 
It contains pretrained full-precision model $P$, quantized model $Q$, and generator $G$.
$G$ is responsible for generating informative and diverse data examples, which are used by a two-level discrepancy function to compute the discrepancy between $P$ and $Q$.
The discrepancy function is composed of output discrepancy $\mathcal{D}_{o}$ and intermediate inter-channel discrepancy $\mathcal{D}_{f}$.
Consequently, $Q$ and $G$ are optimized through a minimax game, where the adversarial learning of the two-level discrepancy modeling is conducted.
In addition, activation regularization $\mathcal{L}_{a}$ encourages $G$ to generate more informative and diverse examples.

In what follows, we first introduce the preliminary of the quantization function used in this paper.
Then we detail the proposed framework.

\subsection{Preliminary}

A common practise in training a neural network with low-precision weights and activations is to introduce a quantization function. Considering the general case of $k$-bit quantization~\cite{zhou2016dorefa}, we define the uniform quantization function $q(\cdot)$ as:
\begin{equation}
  \label{eq:quant}
  q(v) = \mathrm{round}\big( S\cdot(v-Z) \big)\,,
\end{equation}
where $v$ denotes the full-precision (float32) value, $S$ is the scaling factor, and $Z$ is the zero point in float32. According to whether the parameter $Z$ is zero, uniform quantization can be divided into two categories: symmetric quantization and asymmetric quantization.
Here we use symmetric quantization and set $Z=0$. 
Consequently, $S$ is formulated as:
\begin{equation}
  S = \frac{2^{k-1}-1}{\max(|x_{\mathrm{f}}|)}\,,
\end{equation}
where $x_{\mathrm{f}}$ is any one of float32-point numbers.

The key of model quantization is to reduce the discrepancy $\mathcal{D}$ between full-precision model $P$ and low-precision model $Q$ through optimizing $Q$, which can be expressed as:
\begin{equation}\label{eq:quantization}
  Q^* = \min_{Q}\mathcal{D}(P, Q)\,.
\end{equation}

\subsection{Two-level Discrepancy Modeling}

As aforementioned, the framework ZAQ leverages a novel two-level discrepancy function to model the discrepancy between full-precision and quantized models.
First, we assume there is a data example $x_g$ generated by $G$, \ie, $x_g=G(z)$ where $z$ is random noise.
We denote the corresponding prediction outputs for full-precision model $P$ and quantized model $Q$ as $P(x_g)$ and $Q(x_g)$, respectively. There are some distance metrics that can be used to measure the discrepancy, such as Kullback-Leibler (KL) divergence.
KL divergence is efficient in data-driven knowledge transfer or distillation, but it is insufficient to maximize the discrepancy when training generator $G$.
This is because some unexpected samples may be similar in prediction, making the negative KLD too small to optimize.
Instead, we adopt L1 loss to measure the output discrepancy $\mathcal{D}_{o}$ in a more direct way:
\begin{equation}
  \label{eq:D_o}
  \mathcal{D}_{o}(P, Q ; G)=\mathbb{E}_{x_g} \left[\frac{1}{N}\left\| P(x_g)-Q(x_g) \right\|_{1} \right]\,,
\end{equation}
where $N$ is element number in the outputs, for instance, class number for classification and label map size for segmentation. 

\begin{figure}[!t]
  \centering
  \includegraphics[width=0.55\linewidth]{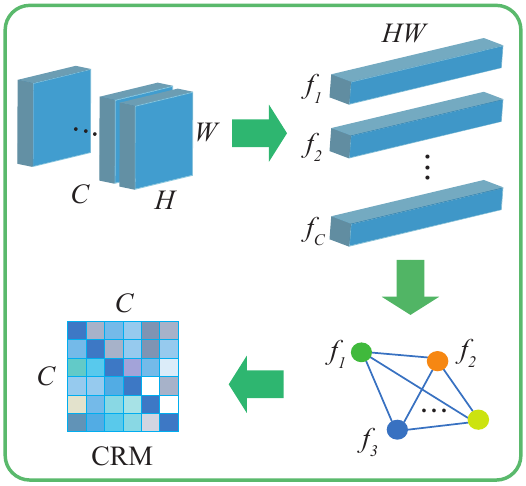}
  \caption{Illustration of obtaining channel relation map.}
  \label{fig:RGM}
\end{figure}

Inspired by the idea of harnessing intermediate feature maps to improve performance in knowledge distillation~\cite{zagoruyko2016paying,park2019relational}, we further propose Channel Relation Map (CRM) to gain intermediate inter-channel discrepancy.
Noting that although there are a few studies~\cite{park2019relational,liu2019knowledge} modeling relations between data instances, we are the first to consider similarity relations between different channels of feature maps, which are introduced later.

Specifically, we define intermediate inter-channel discrepancy as below:
\begin{equation}
  \label{eq:D_f}
  \mathcal{D}_{f}(P,Q;G) = \mathbb{E}_{x_g}\left[\sum_{l}^{L} \frac{\omega^{(l)}}{{C^{(l)}}^2} \left\| \mathcal{R}^{(l)}_{P}(x_g) - \mathcal{R}^{(l)}_{Q}(x_g) \right\|_{1}\right]\,,
\end{equation}
where $L$ is the total number of layers exploited for ZAQ, $\mathcal{R}^{(l)}_{P}(\cdot)$ and $\mathcal{R}^{(l)}_{Q}(\cdot)$ represent CRM extracted from the $l$-th layer of $P$ and $Q$, respectively.
$\omega^{(l)}$ is the adaptive weight allocated to the $l$-th layer, and $C^{(l)}$ is the output channel number of the $l$-th layer. 
Actually, we usually select the last layer in each group or block for residual neural networks, and the layer number is $3\sim4$ for VGG in our experiments.  

A conventional manner to measure the discrepancy of intermediate layers relies on correlating feature maps of $P$ and $Q$, just as what KD commonly does~\cite{RomeroBKCGB14,zagoruyko2016paying}.
However, since the numerical spans of $P$ and $Q$ are very different because of precision settings, the gap between feature maps in $P$ and $Q$ is relatively large (verified in Table~\ref{tab:RGM_ablation}).
Therefore, we introduce CRM to address this issue.
Gram matrix can represent certain relationships between feature vectors to reflect the characteristics of images, and is commonly used in style transfer~\cite{gatys2015neural}. But it is unreliable to measure the feature discrepancy between two networks by directly using feature vectors with different precision. 
Here we extend it to channel relation map to capture the relations towards different channels in the same layer of one model.
It cannot only shield the influence of feature maps with different numerical spans, but also represent the high-dimensional features of samples.
Figure~\ref{fig:RGM} illustrates the procedures of obtaining CRM, which are the same for both $P$ and $Q$.
Taking the feature map $\mathcal{\tilde{F}}^{(l)} \in \mathbb{R}^{C\times H \times W}$ extracted from the $l$-th layer of $P$ (or $Q$) for clarification, it can be flattened into $\mathcal{F}^{(l)} \in \mathbb{R}^{C\times HW}$, which is composited by $C$ channel-wise feature vectors: $\begin{bmatrix} \bm{f}^{(l)}_{1} & \bm{f}^{(l)}_{2} & \cdots & \bm{f}^{(l)}_{C} \end{bmatrix}^{\top}$.
Then the consine similarity between channel features $\bm{f}^{(l)}_i$ and $\bm{f}^{(l)}_j$ is defined as below:
\begin{equation}
\label{eq:rgm}
  \mathcal{R}^{(l)}_{ij}=\frac{ <\bm{f}^{(l)}_i, \bm{f}^{(l)}_j> }{\left\| \bm{f}^{(l)}_i \right\|_2 \left\| \bm{f}^{(l)}_j \right\|_2 } \,.
\end{equation}
Based on $\mathcal{R}^{(l)}_{ij}$ $(i, j\in \{1,2,\dots,C\})$, the corresponding matrices $\mathcal{R}^{(l)}_{P}(x_g)$ and $\mathcal{R}^{(l)}_{Q}(x_g)$ can be obtained. 

To adaptively determine $\omega^{(l)}$ in Eq.~\ref{eq:D_f}, we use the discrepancy calculated for the two models and define the following computational equation:
\begin{smaller}
  \begin{equation}
    \label{eq:adapt_w}
    \omega^{(l)} = \frac{\exp\left(\mathrm{EMA}_{T}\left(\mathbb{E}_{x_g\in\mathcal{B}_t}\left[\left\|\mathcal{R}^{(l)}_{P}(x_g) - \mathcal{R}^{(l)}_{Q}(x_g) \right\|_{1}\right]\right)\right)}{\sum^{L}_{l'}\exp\left(\mathrm{EMA}_{T}\left(\mathbb{E}_{x_g\in\mathcal{B}_t}\left[\left\|\mathcal{R}^{(l')}_{P}(x_g) - \mathcal{R}^{(l')}_{Q}(x_g) \right\|_{1}\right]\right)\right)}\,,
  \end{equation}
\end{smaller}
where $\mathrm{EMA}_{T}$ denotes exponential moving averaging, $T$ is the training steps in an epoch, and $L$ is the number of layers exploited for ZAQ.
By this way, more attention will be paid to the model layer that has a larger difference in CRMs.
Besides, in order to avoid breaking the balance in long-term training, $\omega^{(l)}$ needs to be re-initialized by $\frac{1}{L}$ when a new epoch starts.

\subsection{Adversarial Knowledge Transfer}

Our ZAQ framework trains quantized model $Q$ and generator $G$ in an adversarial minimax game, which contains \textbf{discrepancy estimation} and \textbf{knowledge transfer} stages.
In discrepancy estimation stage, the generator $G$ aims at maximizing the two-level discrepancy between $Q$ and $P$ to search for discrepancy represent space. The loss is defined as follows:
\begin{equation}
  \label{eq:L_DE1}
  \mathcal{L}_{DE} = -\mathcal{D}_{o}(P, Q ; G) - \alpha\mathcal{D}_{f}(P, Q ; G)\,,
\end{equation}
where $\alpha$ is a hyperparameter to balance $\mathcal{D}_{o}$ and $\mathcal{D}_{f}$.

In knowledge transfer stage, quantized model $Q$ is optimized to minimize the two-level discrepancy to approximate full-precision model $P$, denoted as: 
\begin{equation}
  \label{eq:L_KT}
  \mathcal{L}_{KT} = \mathcal{D}_{o}(P, Q ; G) + \alpha\mathcal{D}_{f}(P, Q ; G)\,.
\end{equation}
As a consequence, the knowledge is transferred from $P$ to $Q$ progressively in the zero-shot situation.

\subsection{Activation Regularization}

Although the L1 loss function can relieve the model from falling into some abnormal sample points in discrepancy estimation, they exist all the time and interfere with the generator's exploration of the original input domain. These unexpected samples could make the prediction distributions of the two networks consistent but they are not in the working domain of full-precision model. 
We assume the infinite discrepancy space between model $P$ and $Q$ is $\mathbf{\Omega}$, in which the generator $G$ explore valuable samples for transfer learning. In fact, $\mathbf{\Omega}$ consists of two subspaces $\mathbf{\Omega}_P$ and $\mathbf{\Omega}_U$, which means $\mathbf{\Omega} = \mathbf{\Omega}_P \cup \mathbf{\Omega}_U$. $\mathbf{\Omega}_P$ is the subspace that is equal to the original training data domain, or the working domain of pretrained model $P$. And $\mathbf{\Omega}_U$ is an infinite subspace outside the working domain of $P$. 
The goal of the generator is to synthesize samples distributed in the subspace $\mathbf{\Omega}_P$, rather in $\mathbf{\Omega}_U$. 

According to several researches about interpretability of DNNs~\cite{zeiler2014visualizing,dong2017towards} or sample reconstruction~\cite{lopes2017data,chen2019data}, the activation layer reflects the sensitivity of the neural network to the input data, and higher activation means more correlation between synthetic samples and working domain of $P$. 
Hence, we further leverage activation regularization to constraint the generator to explore and synthesize valuable samples. We denote the $i$-th channel activation map extracted by the last convolution layer of network $P$ as $h^P_i, i\in \{1,2,\dots,M\}$, where $M$ is the number of activation maps. Then, the activation regularization can be formulated as 

\begin{equation}
  \label{eq:L_a}
  \mathcal{L}_{a} = -\frac{1}{M}\sum_{i}^{M}\left \| h^{P}_{i} \right \|_{1}\,.
\end{equation}

With the intuition that high activation values mean a better matching between a given input example and training data, we incorporate $\mathcal{L}_{a}$ into Eq.~\ref{eq:L_DE1} and minimize the following loss to guide generator training.
\begin{equation}
  \label{eq:L_DE}
  \mathcal{L}_{DE} = -\mathcal{D}_{o}(P, Q ; G) - \alpha\mathcal{D}_{f}(P, Q ; G) + \beta\mathcal{L}_{a}.
\end{equation}

Finally, the detailed procedures of the proposed framework ZAQ is summarized in Algorithm~\ref{alg:ZAQ}.

\begin{algorithm}[ht]
  \caption{Zero-shot Adversarial Quantization}
  \label{alg:ZAQ}
  \SetAlgoLined
  \LinesNumbered
  \KwIn{A pretrained full-precision model $P(x;\theta^p)$, quantization precision.}
  \KwOut{Quantized model $Q(x;\theta^q)$}
  Quantize the model $P$ as $Q$ by Eq.~\ref{eq:quantization}\;

  \For{number of epochs}{
    Initialize adaptive weights by $\frac{1}{L}$\;
    \For{number of training steps}{
      \textbf{\# Discrepancy Estimation}\\
      $z \sim \mathcal{N}(\mathbf{0},\boldsymbol{I})$, $x_g \leftarrow G(z;\theta^g)$\;
      Estimate $\mathcal{L}_{DE}$ by Eq.~\ref{eq:L_DE}\;
      Fix $\theta^q$, update $\theta^g$: 
      $$ \theta^g \leftarrow \theta^g - \eta\frac{\partial\mathcal{L}_{DE}}{\partial\theta^g} $$
      Update adaptive weights $\omega^{(l)}$ by Eq.~\ref{eq:adapt_w}\;

      \textbf{\# Knowledge Transfer}\\
      $z \sim \mathcal{N}(\mathbf{0},\boldsymbol{I})$, $x_g \leftarrow G(z;\theta^g)$\;
      Calculate $\mathcal{L}_{KT}$ by Eq.~\ref{eq:L_KT}\;
      Fix $\theta^g$, update $\theta^q$:
      $$ \theta^q \leftarrow \theta^q - \eta\frac{\partial\mathcal{L}_{KT}}{\partial\theta^q} $$
    }
    decay $\eta$\;
  }
\end{algorithm}

\section{Experiments}


\begin{table*}[!t]
  \centering
  \resizebox{\linewidth}{!}{
    \begin{tabular}{c|cc|ccc|cccccc}
    \toprule
    Dataset & \multicolumn{1}{c}{Model} & size (MB) & \multicolumn{1}{c}{bit} & size (MB) & float32 & RQ & ZeroQ & GDFQ  & DFQ   & ACIQ  & ZAQ \\
    \midrule
    \multirow{2}[2]{*}{CIFAR10} & MobileNetV2 & 9.0   & W6A6  & 1.7   & 92.39 & 78.90 & 89.90 & 91.27 & 85.43 & 91.04  & \textbf{92.15} \\
          & VGG19 & 149   & W4A8  & 25.1  & 93.49 & 92.42 & 92.69 & 92.84 & 92.66 & 92.48  & \textbf{93.06} \\
    \midrule
    \multirow{2}[2]{*}{CIFAR100} & ResNet20 & 1.1   & W5A5  & 0.2   & 69.58 & 49.54 & 65.7  & 66.12 & 59.42 & 60.19  & \textbf{67.94} \\
          & ResNet18 & 43    & W4A4  & 5.4   & 77.38 & 17.00 & 70.25 & 71.53 & 40.35 & 54.73 & \textbf{72.67} \\
    \midrule
    \multirow{3}[2]{*}{ImageNet} & MobileNetV2 & 14    & W8A8  & 3.5   & 71.88 & 67.09 & 70.88 & 70.17 & 70.58 & 68.92 & \textbf{71.43} \\
          & \multicolumn{1}{c}{ResNet50} & 98    & \multicolumn{1}{c}{W4A4} & 12.3  & 76.13 & 64.90  & 69.30  & 68.69 & 10.32 & 59.34 & \textbf{70.06} \\
          & ResNet50 & 98    & W2A2  & 6.1   & 76.13 & 11.25 & 63.12 & 64.96 & 1.48  & 3.25  & \textbf{65.52} \\
    \bottomrule
    \end{tabular}%
  }
  \caption{Results of image classification on three datasets.}
  \label{tab:classification}%
\end{table*}%

\subsection{Experimental Setup}

\subsubsection{Datasets}
We evaluate our approach on the following six datasets: CIFAR10, CIFAR100, and ImageNet for classification, Cityscapes and CamVid for segmentation, and VOC2012 for object detection. 

\textbf{CIFAR.} CIFAR10~\cite{krizhevsky2009learning} and CIFAR100 consist of 32$\times$32 color images with 10 and 100 classes, respectively. Both are split into a 50,000-image train set and a 10,000-image test set.

\textbf{ImageNet.} The 1,000-class dataset from ILSVRC 2012~\cite{ILSVRC15} provides 1.2 million images for training, and 50,000 for validation.

\textbf{Cityscapes.} Cityscapes~\cite{Cordts2016Cityscapes} is for urban scene understanding and contains 30 classes with only 19 classes used for evaluation.
It provides 3,975 images with fine segmentation annotations, including 2,975 images for training and 500 images for testing.

\textbf{CamVid.} CamVid~\cite{brostow2008segmentation} is an automotive dataset, containing 367 training and 233 testing images. We perform on the commonly used 11 different classes.

\textbf{VOC2012.} A total of 11540 images are included in PASCAL VOC2012~\cite{everingham2011pascal}, where each image contains a set of objects, out of 20 different classes. 

\subsubsection{Baselines} 
To evaluate the effectiveness and advantages of our proposed method, we compared it with both data-free fine-tuning methods and post-training quantization methods. The baselines are briefly described as follows.

\textbf{FT.} We use original training data to Fine-Tune~(FT) a quantized model. 

\textbf{RQ.} Raw Quantization~(RQ) method directly testing the model after quantization without any fine-tuning.

\textbf{DFQ~\cite{nagel2019data}.} A post-training quantization method uses a weight equalization scheme to remove outliers in both weights and activations.

\textbf{ACIQ~\cite{banner2018aciq}.} It analytically computes a clipping range, as well as a per-channel bit allocation for neural networks without any fine-tuning/training.

\textbf{ZeroQ~\cite{cai2020zeroq}.} It retrains a quantized model by reconstructed data instead of original data. 

\textbf{GDFQ~\cite{xu2020generative}.} It is also a fine-tuning method by recovering fake data via a conditional generator. Yet it only supports classification tasks.

\subsubsection{Implementation Details}
We implement all networks and quantization methods in Pytorch. For all datasets, we adopt the same data augmentation procedure on pretraining as~\cite{sandler2018mobilenetv2} for making fair comparisons. We adopt SGD with momentum 0.9 and weight decay $5\times10^{-4}$ in both pretraining and fine-tuning. All the models are pretrained for 200 epochs and the learning rates are decayed by 0.1 for every 80 epochs on datasets, except ImageNet, on which we directly use the official pretrained models. We construct a generator following DCGAN~\cite{radford2015unsupervised} with 256-dimension noise and is trained with Adam~\cite{kingma2014adam}. But for CIFAR, we just reduce the channels of all layers in the generator to a quarter and set the dimension of noise to 100, due to the smaller size of the samples. Moreover, the learning rates of quantized models and generators are initialized to 0.1 and $1\times10^{-3}$, respectively. The learning rates of SGD and Adam are decayed by different steps in different tasks. In training, we set the batch size to 256 for CIFAR, 64 for ImageNet and VOC2012, and 16 for segmentation datasets. As for the hyperparameters, we set $\alpha=0.1$ and $\beta=0.05$ by default. More detailed implementation and settings for different datasets are illustrated in the following parts.

\subsection{Experimental Results}

\subsubsection{Performance Test for Image Classification} For image classification, we take the top-1 accuracy~(abbr. Acc) as the metric. 
The number of fine-tuning epochs is 200 for CIFAR, while 300 for ImageNet.
In each epoch, the training steps are set to 40 for CIFAR and 50 for ImageNet. 
The learning rates of SGD and Adam are decayed every 80 epochs for CIFAR and 100 for ImageNet.
Besides, we use ``W-A-'' to denote the quantization bits used for weights~(W) and activations~(A), and ``float32'' as full-precision models.

Table~\ref{tab:classification} shows the classification results.
First, we find DFQ and ACIQ suffer from dramatic performance degradation when taking ultra-low precision, especially for CIFAR100 and ImageNet.
This verifies that due to the lack of fine-tuning, post-training quantization methods do not work well for ultra-low precision.
Then we observe our framework achieves the best performance on the three classification datasets, indicating its advantages over the other quantization methods.

\begin{figure}[!t]
  \centering
  \includegraphics[width=\linewidth]{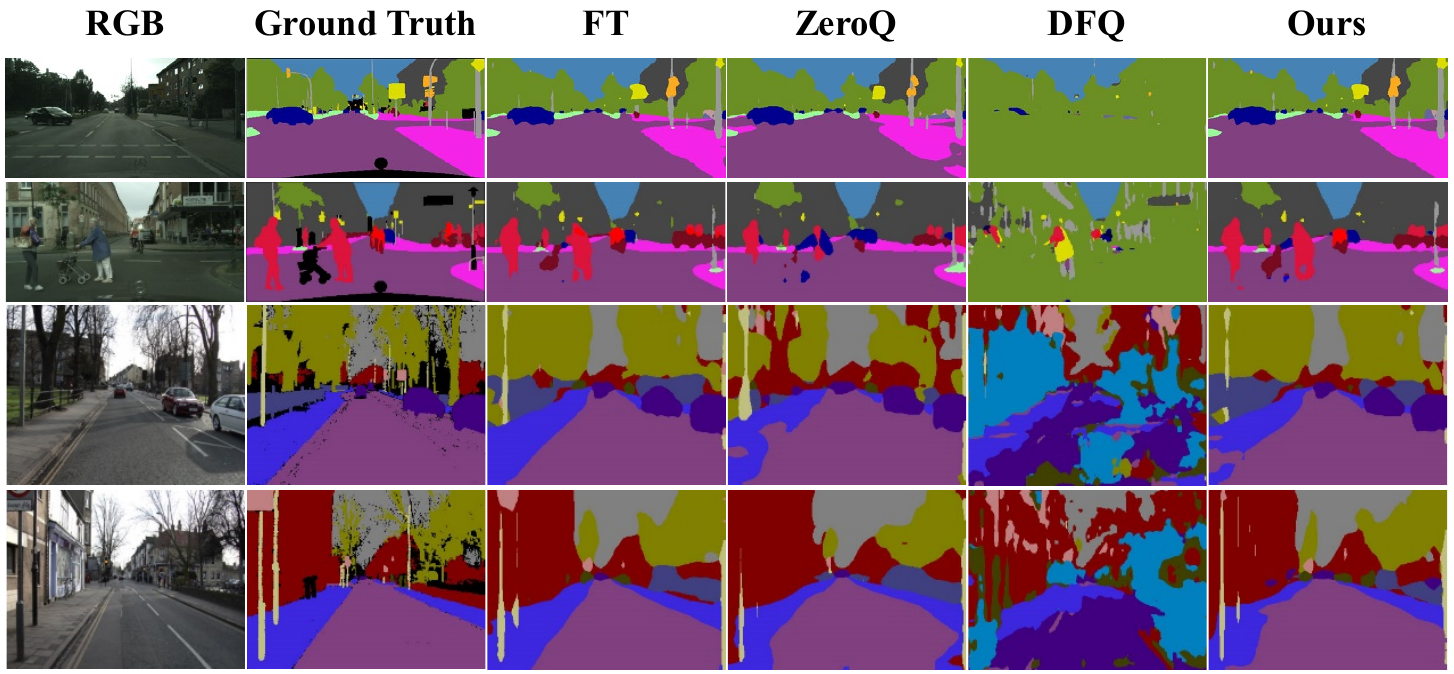}
  \caption{Visualization of segmentation results for Cityscapes (the first two rows) and CamVid (the last two rows).}
  \label{fig:visual_real_seg}
\end{figure}

\subsubsection{Performance Test for Image Segmentation}

In this part, we mainly compare ZAQ with ZeroQ and DFQ on Cityscapes and CamVid, the images of which are all resized to 256.
GDFQ requires labels as conditions to synthesize data, so it does not naturally support high-level vision tasks such as segmentation and detection.
The ImageNet-pretrained MobileNetV2 and ResNet50 models are used as feature extractors within DeepLabv3~\cite{chen2017rethinking}. The hyperparameters $\alpha=0.5$ and $\beta=0.1$. 
We adopt mean IoU of all classes~(mIoU) as the evaluation metric for segmentation. 
In fine-tuning, we set the size of the synthetic image as $128\times128$, which is enough for representing model discrepancy and transferring knowledge.

Table~\ref{tab:seg} shows the performance of quantized models fine-tuned by different methods, from which we can see that our method still exhibits superior performance, especially for ultra-low precision situations.
This observation is consistent with what we find in image classification tasks.

Furthermore, we randomly select two real examples from Cityscapes and CamVid, respectively, and visualize the segmentation results of 4-bit DeeplabV3(MobileNetV2) learned by different model quantization methods.
The results are shown in Figure~\ref{fig:visual_real_seg}, where the first two rows correspond to Cityscapes and the last two rows correspond to CamVid.
Obviously, DFQ does not work properly for the examples in 4-bit quantization and thus it is hard to retain model performance. 
By comparing ZAQ with ZeroQ, we find it exhibits better qualitative results in complex details and small object segmentation, as shown in the second row of the figure.

\begin{table}[!t]
  \centering
  \resizebox{\linewidth}{!}{
    \begin{tabular}{c|c|cccc}
    \toprule
    Dataset & Method & W8A8  & W6A6  & W4A4  & W2A2 \\
    \midrule
    \multirow{5}{*}{\shortstack{Cityscapes\\(63.39)}}
    & FT    & 61.25 & 59.64 & 55.98 & 45.77 \\\cline{2-6}    
    & RQ & 58.42 & 55.33 & 29.16 & 0.44 \\
          & DFQ   & 57.34 & 55.29 & 19.06 & 3.13 \\
          & ZeroQ & 59.52 & 57.97 & 52.73 & 43.18 \\
          & ZAQ   & \textbf{60.18} & \textbf{58.12} & \textbf{55.12} & \textbf{44.93} \\
    \midrule
    \multirow{5}{*}{\shortstack{CamVid\\(53.34)}} 
          & FT    & 52.76 & 50.75 & 49.13 & 40.06 \\\cline{2-6}
         & RQ & 44.96 & 43.20  & 10.05 & 0.02 \\
          & DFQ   & \textbf{51.02} & 46.13 & 11.78 & 2.33 \\
          & ZeroQ & 49.92 & 48.56 & 43.83 & 36.44 \\
          & ZAQ   & 50.89 & \textbf{49.77} & \textbf{47.62} & \textbf{39.95} \\
    \bottomrule
    \end{tabular}%
  }
  \caption{Results on Cityscapes and CamVid (mIoU).}
  \label{tab:seg}
\end{table}%

\subsubsection{Performance Test for Object Detection}

To demonstrate the application on object detection, we apply ZAQ to the model MobileNetV2 SSD~\cite{liu2016ssd} and evaluate it on VOC2012. Table~\ref{tab:obj_det} briefly demonstrates the advantages of our method compared to other quantization methods. 
In particular, ZAQ is comparable with FT that utilizes the original training dataset.

\begin{table}[t]
  \centering
  \begin{tabular}{c|cccc}
  \toprule
  Method & W8A8 & W4A8 & W4A4 & W2A2 \\
  \midrule
  FT    & 70.35 &	68.24 & 64.28 &	57.02 \\\hline
  RQ    & 68.31 & 66.25 & 5.27 & 1.06 \\
  DFQ   & 69.16 & 64.57 &	13.15 & 2.65 \\
  ZeroQ & 69.04 & 67.53 &	62.72 & 53.07 \\
  ZAQ   & \textbf{70.02}& \textbf{68.12}&	\textbf{64.44}& \textbf{56.96} \\
  \bottomrule
  \end{tabular}
  \caption{Results of SSD(MobileNetV2) on VOC2012 (mAP).}
  \label{tab:obj_det}
  \vspace{-2mm}
\end{table}

\begin{figure}[!h]
  \vspace{-2mm}
  \centering

  \subfigure[ResNet18 on ImageNet]{
  \includegraphics[width=0.43\linewidth]{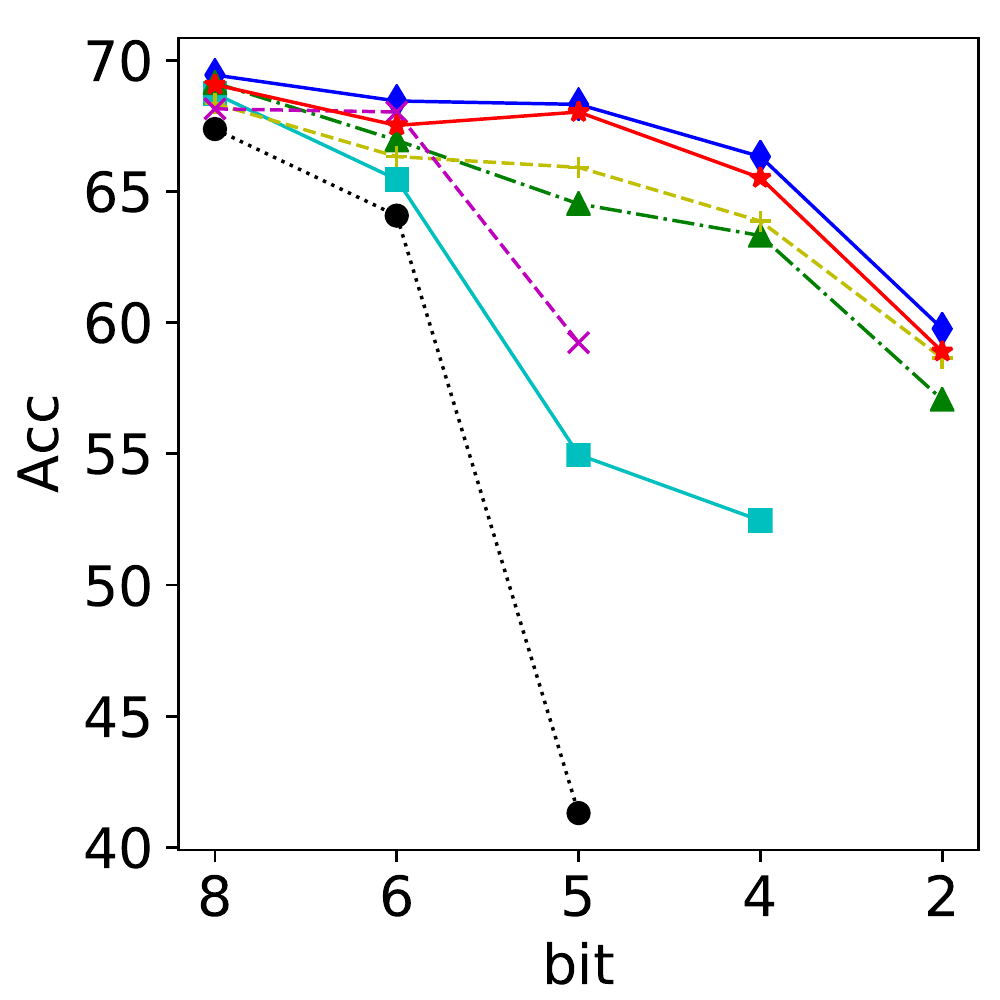}
  }
  \subfigure[DeeplabV3 on Cityscapes]{
  \includegraphics[width=0.43\linewidth]{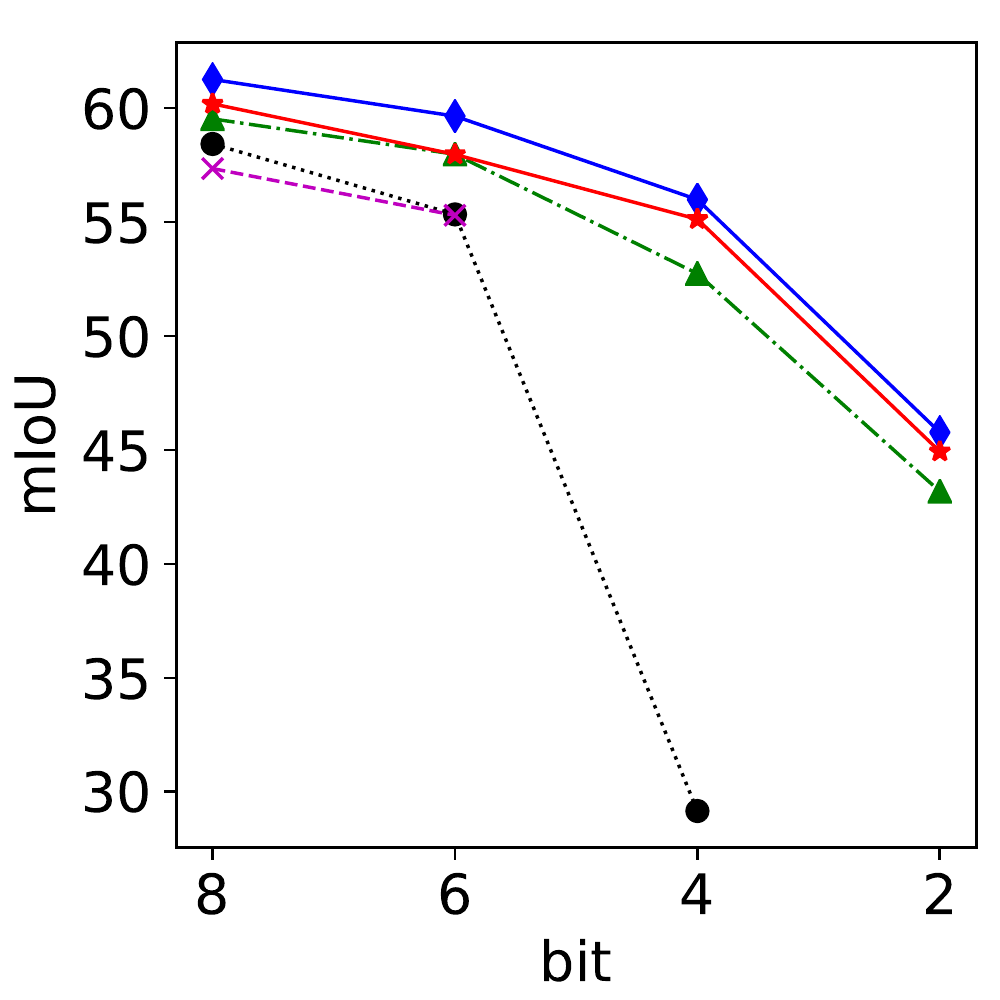}
  }

  \subfigure[SSD on VOC2012]{
  \includegraphics[width=0.43\linewidth]{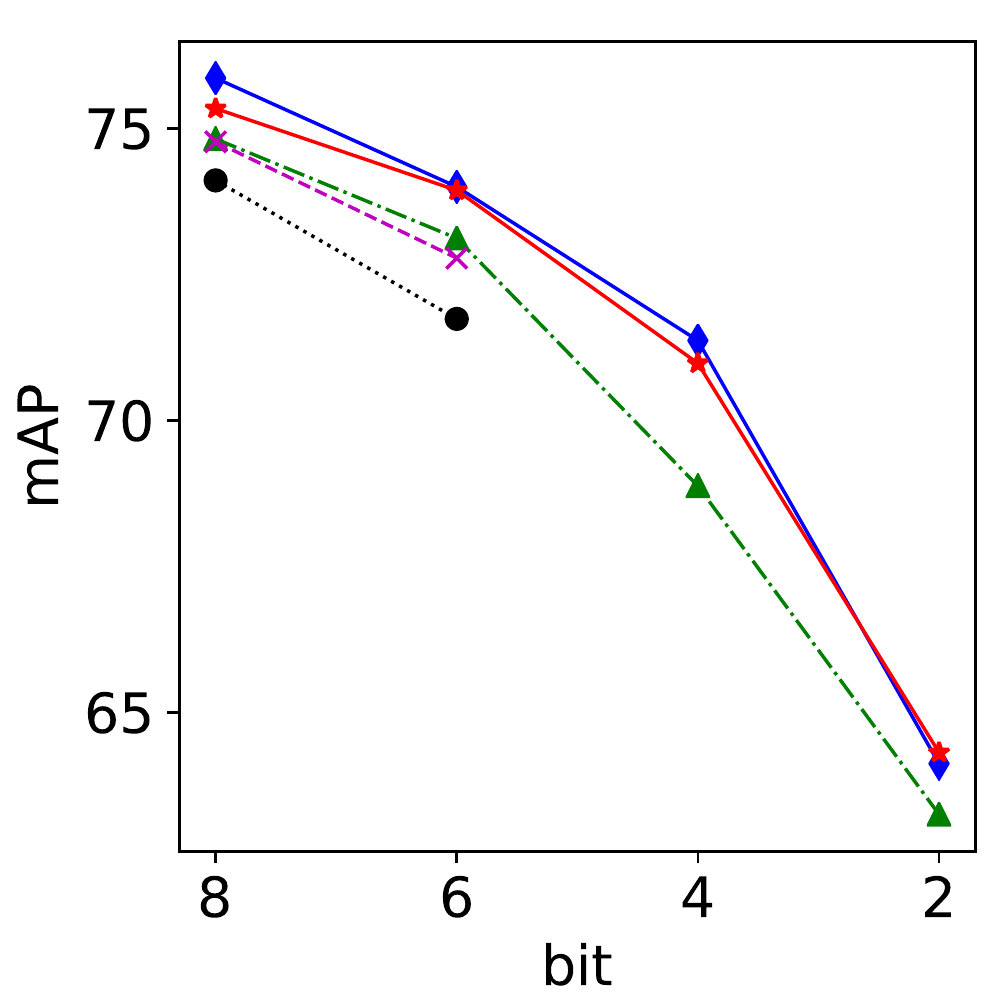}
  }
  \subfigure{
  \centering
  \includegraphics[width=0.43\linewidth]{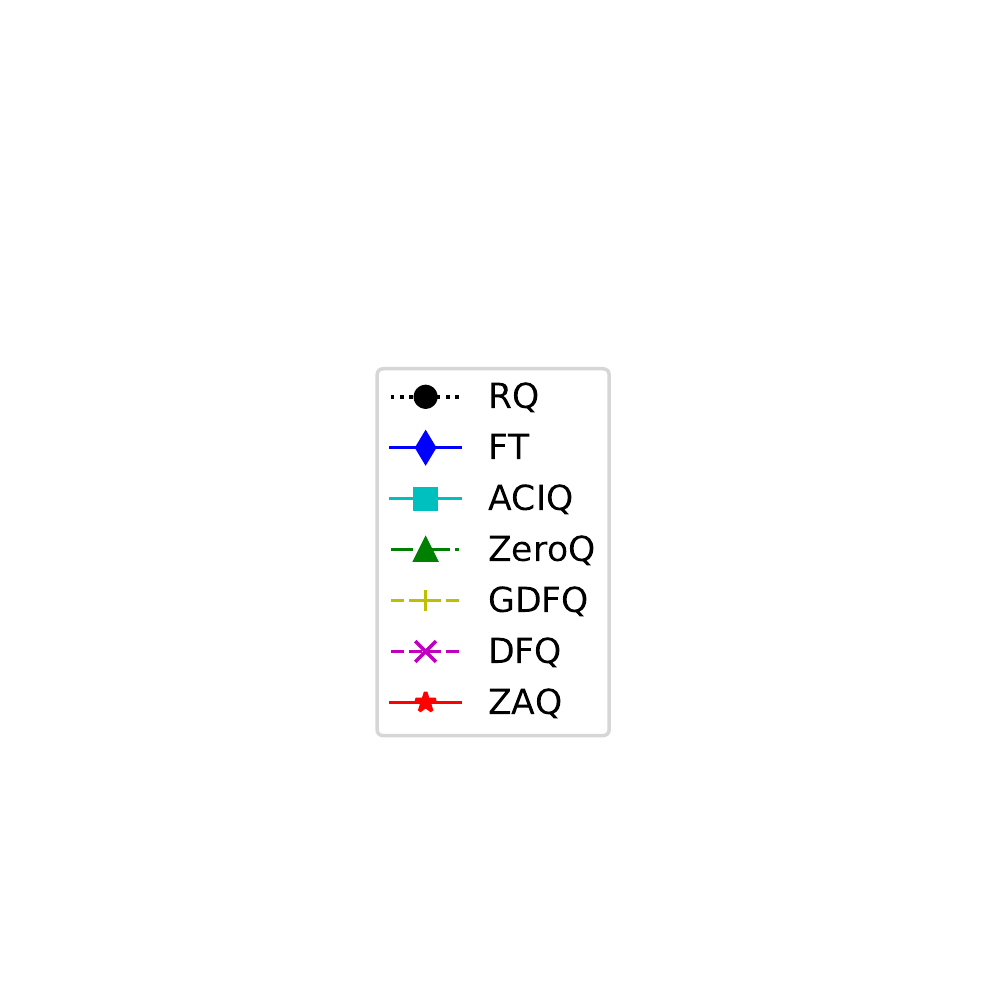}
  }
  \caption{Performance change versus different quantization precision.}
  \label{fig:bit_Acc}
  \vspace{-2mm}
\end{figure}

Finally, we end up the introduction of the performance tests for three image-based tasks with Figure~\ref{fig:bit_Acc}, which provides an overview of how performance changes with different bits.
The curves of different quantization methods in the figures reflect that ZAQ is consistently better and it gains greater improvements in ultra-low precision situation. 

\subsubsection{Ablation Study}

\begin{figure}[!t]
  \centering
  \subfigure[ImageNet]{
  \includegraphics[width=0.47\linewidth]{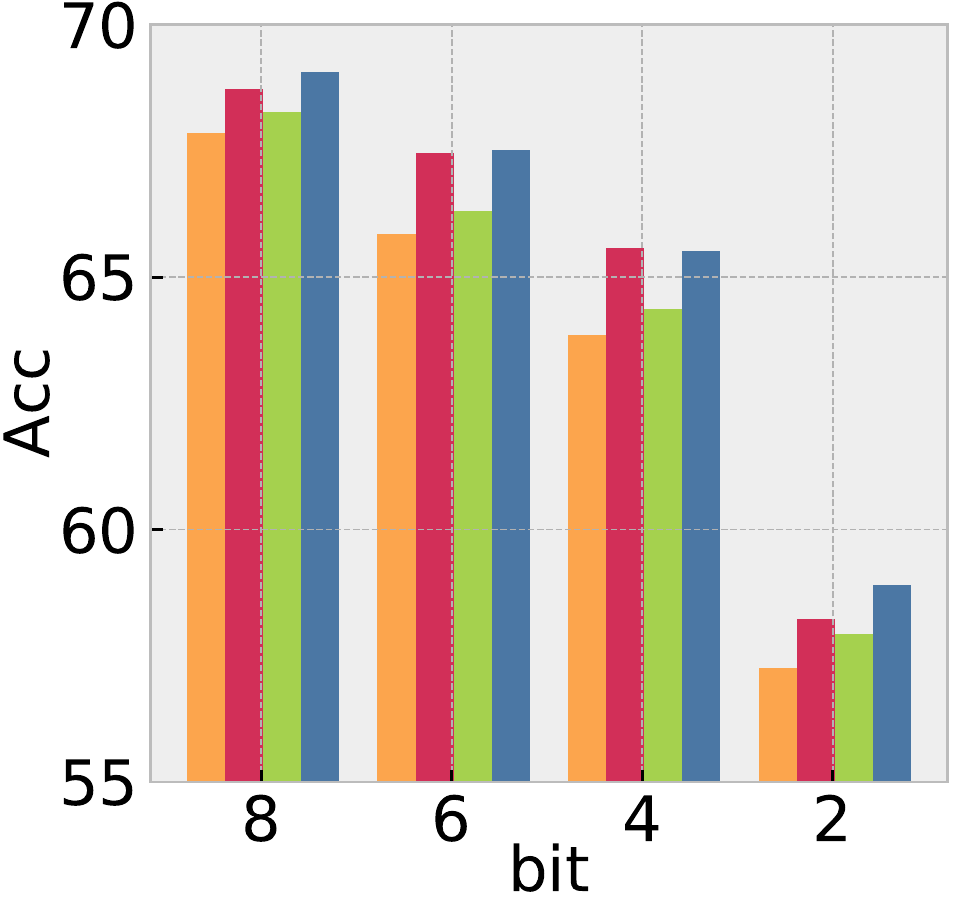}
  }
  \subfigure[Cityscapes]{
  \includegraphics[width=0.47\linewidth]{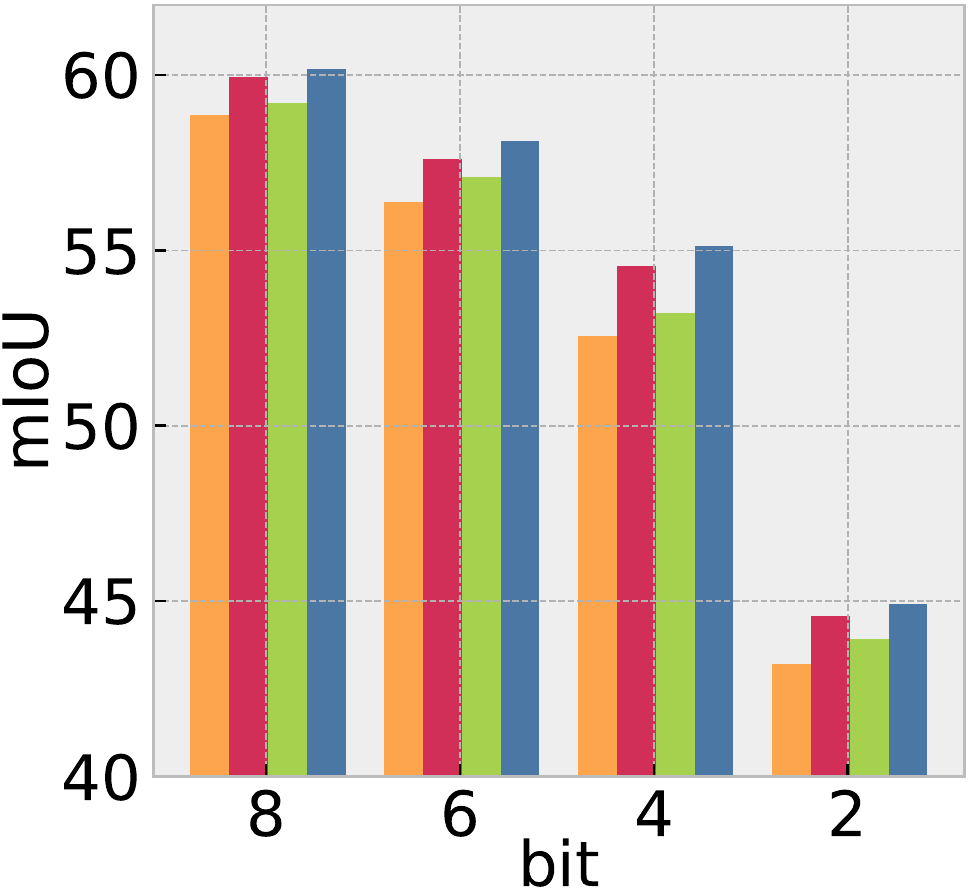}
  }
  \subfigure{
  \includegraphics[width=0.85\linewidth]{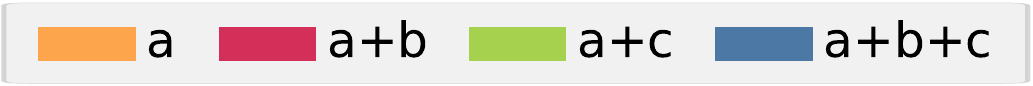}
  }
  \caption{Effectiveness of different components of the proposed ZAQ method. Noting that `a':$\mathcal{D}_o$, `b':$\mathcal{D}_f$, `c':$\mathcal{L}_a$.}
  \label{fig:ablation}
  \vspace{-2mm}
\end{figure}

In this part, we conduct an ablation study to validate the contributions of the main components in ZAQ.
First of all, Figure~\ref{fig:ablation} presents the benefits of output discrepancy $\mathcal{D}_o$~(`a'), intermediate inter-channel discrepancy $\mathcal{D}_f$~(`b'), and activation regularization $\mathcal{L}_a$~(`c') on ImageNet (using model ResNet18) and Cityscapes (using model DeeplabV3(ResNet50)).
Since output discrepancy $\mathcal{D}_o$ is directly associated with the final model output, it should not be removed anytime.
As we can see, the intermediate inter-channel discrepancy could bring $1\sim2\%$ performance improvement, while the activation regularization has a smaller contribution of about 0.5\% improvement. 
But it could prevent the generator from falling into some abnormal samples that are not sensitive to the full-precision models, which is also utilized in the previous study for KD~\cite{chen2019data}.

\begin{table}[ht]
  \centering
  \resizebox{\linewidth}{!}{
    \begin{tabular}{c|c|c|c|c|c}
    \toprule
    Dataset & Model & bit   & CRM   & Gram    & AT \\
    \midrule
    CIFAR100 & ResNet18 & W4A4  & 72.67 & 45.32 & 61.80 \\
    Cityscapes & DeeplabV3 & W8A8  & 52.17 & 41.36 & 48.65 \\
    \bottomrule
    \end{tabular}%
  }
  \caption{Ablation study of CRM.}
  \label{tab:RGM_ablation}%
\end{table}%

\begin{figure}[b]
  \vspace{-2mm}
	\centering
	\subfigure[full-precision model]{
		\label{fig:P_RGM}
		\includegraphics[width=0.4\linewidth]{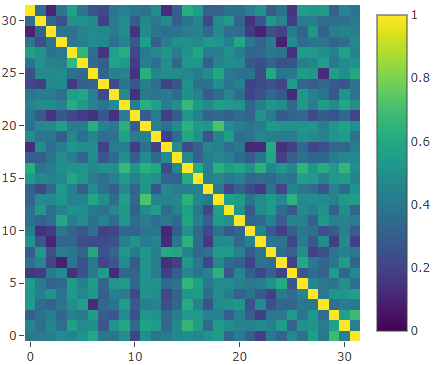}
	}
	\subfigure[quantization model]{
		\label{fig:Q_RGM}
		\includegraphics[width=0.4\linewidth]{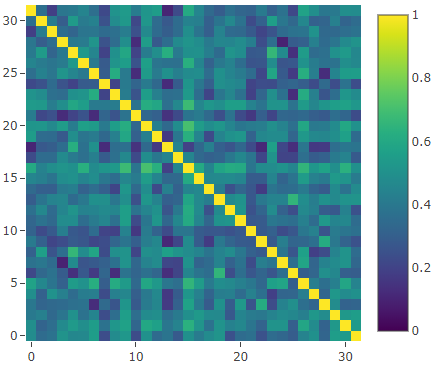}
	}
	\caption{Visualization of CRMs on CIFAR100. }
  \label{fig:visual_RGM}
\end{figure}

We further demonstrate the effectiveness of CRM in model quantization by comparing it with two alternatives for learning intermediate knowledge: (1) Gram which directly uses Gram matrix in discrepancy modeling; (2) AT~\cite{zagoruyko2016paying} which directly aligns normalized feature maps in knowledge transfer.
Table~\ref{tab:RGM_ablation} shows the performance of the above-mentioned methods, from which we can see CRM is much better than the other two methods.
This verifies the necessity of considering different numerical spans in designing quantization-aware fine-tuning methods.
In addition, we choose CIFAR100 to visualize the computed CRMs by ZAQ.
In Figure~\ref{fig:visual_RGM}, (a) and (b) are CRMs from the 2-nd exploited layer of full-precision and 4-bit ResNet18, respectively.
By comparison, we can find the two CRMs are consistent with each other.

\subsubsection{Efficiency Analysis}

We conduct efficiency test on a single GPU (GTX 2080Ti) for ZAQ and the data generation-based quantization methods, \ie, ZeroQ and GDFQ.
The number of synthesized images determined for each method is conditioned on its performance convergence state following \cite{cai2020zeroq} and \cite{xu2020generative}. 
Due to the poor diversity of synthetic images, ZeroQ and GDFQ need to synthesize more samples in training. 
Besides, the images in Cityscapes have high resolution, making ZeroQ cost too much time in synthesizing procedure.
So we conduct the comparative experiment on Cityscapes with the same number of samples approximate to the original dataset in each epoch.
Table~\ref{tab:performance} shows the results, where our method reduces GPU time by 41.8\% compared to GDFQ on CIFAR100, while 57.5\% compared to ZeroQ on Cityscapes.
The conclusion is intuitive since ZeroQ needs 500 to 1500 iterations to generate per image and GDFQ is prone to generate redundant images.

\begin{table}[!h]
  \centering
  \resizebox{0.75\linewidth}{!}{
    \begin{tabular}{c|cc|c}
    \toprule
    Dataset & Method & images & GPU time \\
    \midrule
    \multirow{3}[2]{*}{CIFAR100} & ZeroQ & 10000 & 5.5 h \\
          & GDFQ  & 12800 & 7.6 h \\
          & Ours  & 5120  & 3.2 h \\
    \midrule
    \multirow{2}[2]{*}{Cityscpaes} & ZeroQ & 1280  & 12.7 h \\
          & Ours  & 1280  & 5.4 h \\
    \bottomrule
    \end{tabular}%
  }
  \caption{Time cost comparison.}
  \label{tab:performance}%
  \vspace{-2mm}
\end{table}%

\subsection{Case Study of Generated Images}

\begin{figure}[!t]
  \centering
  \includegraphics[width=\linewidth]{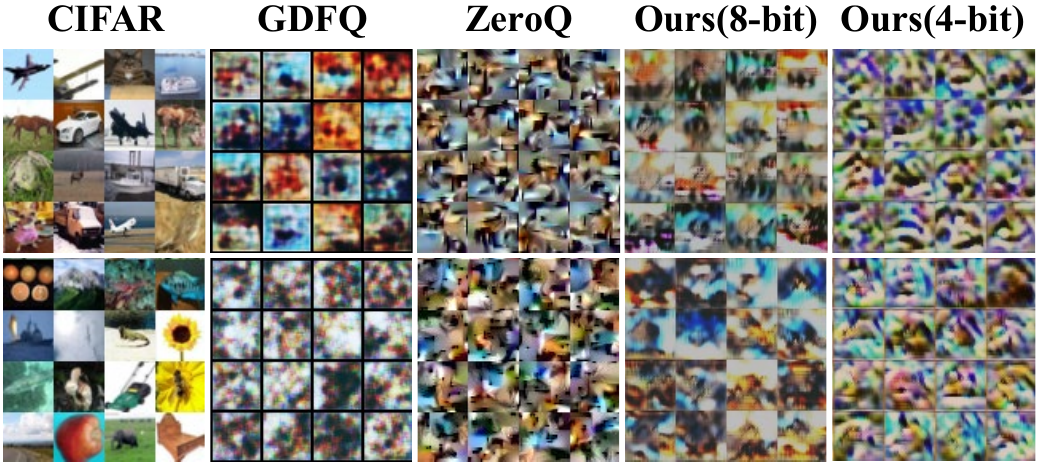}
  \caption{Generated samples about CIFAR.}
  \label{fig:cifar_gen}
\end{figure}

This part conducts case studies on the generated data by different model quantization methods.
We take CIFAR and CamVid for illustration.
For CIFAR, the randomly selected images are shown in Figure~\ref{fig:cifar_gen}.
The first row of the images corresponds to CIFAR10 and the second row corresponds to CIFAR100.
The first column shows the original images. 
The images in the middle three columns are gotten from MobileNetV2 (quantized to 8 bits) and the last column is from ResNet20 (quantized to 4 bits).
By investigating the image patterns generated by GDFQ and ZeroQ, we find there is a big gap between them and those of the original images.

Although the image samples by ZAQ seem to be not recognizable by humans or be similar to the original data, their goal is to represent the discrepancy between two models with different precision.
The comparison between the synthetic images of ZQA and those of GDFQ and ZeroQ indicates that ZQA could generate more diverse images, while GDFQ and ZeroQ suffer from more repeated patterns in their generated images.
This empirically shows the efficiency of knowledge transfer in ZAQ.

\begin{figure}[!t]
  \centering
  \includegraphics[width=0.7\linewidth]{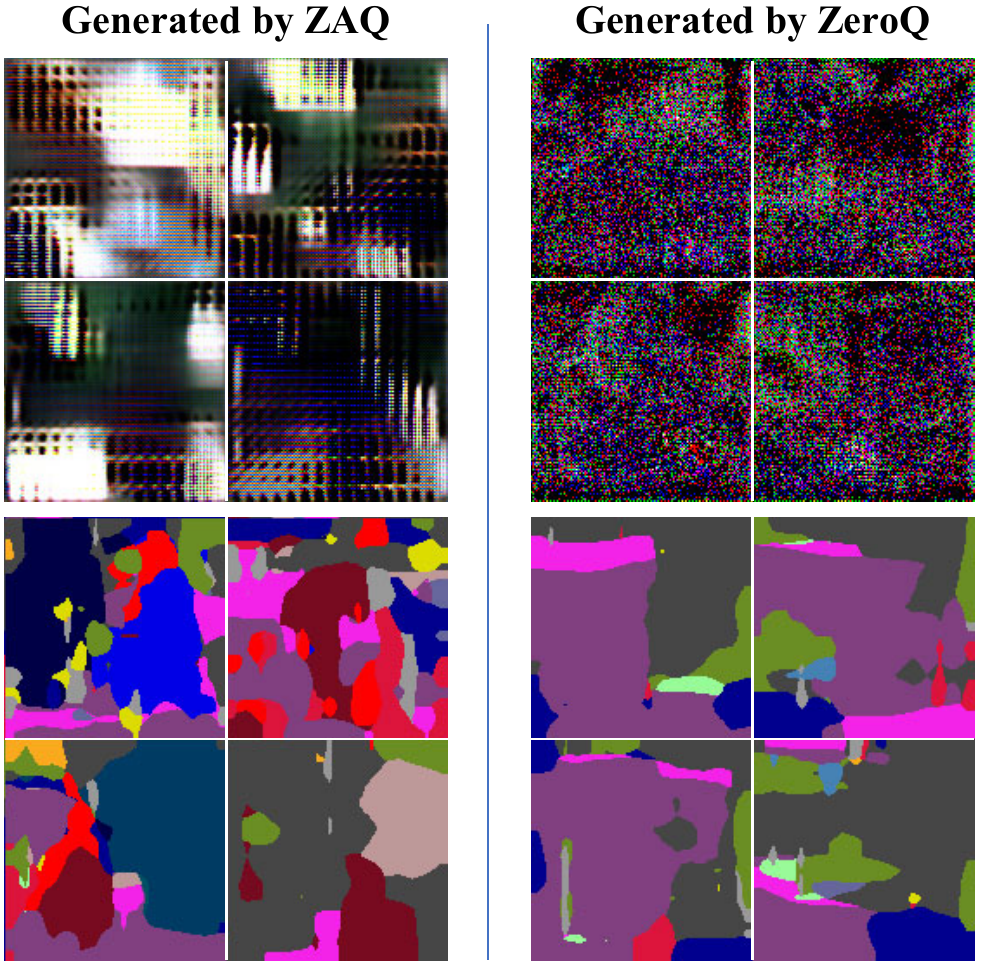}
  \caption{Sematic images about CamVid generated by ZAQ and ZeroQ based on DeeplabV3(MobileNetV2).}
  \label{fig:seg_gen}
\end{figure}

Furthermore, we visualize the semantic image samples generated by ZAQ and ZeroQ in Figure~\ref{fig:seg_gen}.
The observation is accordant with what we have observed in image classification datasets.
That is, ZAQ tends to generate semantic images with more diversity, while ZeroQ reconstructs semantic images with some duplicated local patterns. 

\section{Conclusion and Future Work}

In this paper, we have proposed ZAQ, a novel zero-shot adversarial quantization framework without needing to access any original training data.
Its main innovations lie in applying adversarial learning to data-free model quantization through alternating two-level discrepancy estimation and knowledge transfer. Our framework is welcomed for its ability of modeling prediction discrepancy, as well as intermediate inter-channel discrepancy between full-precision and quantized models. 
Extensive experiments on various deep neural models for three common vision tasks demonstrate the superiority of ZAQ, especially for ultra-low precision situations.
In the future work, we consider applying the proposed method to other domains such as BERT quantization~\cite{shen2020q}, and extending ZAQ to automatic mixed precision quantization.

\vspace{-2mm}
\paragraph{Acknowledgement:} This work was supported in part by National Natural Science Foundation of China under Grant (No. 62072182).

{\small
\bibliographystyle{ieee_fullname}
\bibliography{ref}
}

\end{document}